\documentclass[runningheads]{llncs}
\usepackage[T1]{fontenc}
\usepackage{xcolor,soul}
\usepackage{graphicx}
\begin{document}
\title{Visual Homing in Outdoor Robots Using Mushroom Body Circuits and Learning Walks}

% Learning Walk and mushroom bodies mechanism for Outdoor Visual Homing 

% Outdoor Visual Homing with Learning Walks and Mushroom Body Circuits

% Ant-Like Visual Homing with Mushroom Bodies in the Wild

% Outdoor visual homing with ant-inspired Mushroom Bodies categorized by the path integrator

% Visual Homing for Outdoor Robots Using Mushroom Bodies Trained During Learning Walks

%Visual Homing for Outdoor Robots via Mushroom Bodies learning guided by Learning Walks

%Visual Homing for outdoor robots using ant’s mushroom-bodies circuits and learning walks

% Learning walks 

\titlerunning{Visual Homing in Outdoor Robots}

%Autonomous Outdoor Visual Homing via a Lateralized Mushroom Body Model and PI Scaffold

\author{Gabriel G. Gattaux\inst{1}\orcidID{0000-0002-9424-7543} \and Julien R. Serres\inst{1,2}\orcidID{0000-0002-2840-7932} \and Franck Ruffier\inst{1}\orcidID{0000-0002-7854-1275} \and Antoine Wystrach\inst{3}\orcidID{0000-0002-3273-7483}}
\authorrunning{G. G. Gattaux et al.}
% First names are abbreviated in the running head.
% If there are more than two authors, 'et al.' is used.
%
\institute{Aix Marseille Univ, CNRS, ISM, Marseille,  France \and Institut Universitaire de France, IUF, Paris,  France \email{\{gabriel.gattaux,julien.serres,franck.ruffier\}@univ-amu.fr}\\ \and  Univ Toulouse, CRCA, CBI, UMR CNRS-UPS 5169, Toulouse, France\\ \email{antoine.wystrach@univ-tlse3.fr}}
\maketitle              % typeset the header of the contribution
\begin{abstract}
Ants achieve robust visual homing with minimal sensory input and only a few learning walks, inspiring biomimetic solutions for autonomous navigation. While Mushroom Body (MB) models have been used in robotic route following, they have not yet been applied to visual homing. We present the first real-world implementation of a lateralized MB architecture for visual homing onboard a compact autonomous car-like robot. We test whether the sign of the angular path integration (PI) signal can categorize panoramic views, acquired during learning walks and encoded in the MB, into “goal on the left” and “goal on the right” memory banks, enabling robust homing in natural outdoor settings. We validate this approach through four incremental experiments: (1) simulation showing attractor-like nest dynamics; (2) real-world homing after decoupled learning walks, producing nest search behavior; (3) homing after random walks using noisy  PI emulated with GPS-RTK; and (4) precise stopping-at-the-goal behavior enabled by a fifth MB Output Neuron (MBON) encoding goal-views to control velocity. This mimics the accurate homing behavior of ants and functionally resembles waypoint-based position control in robotics, despite relying solely on visual input. Operating at 8 Hz on a Raspberry Pi 4 with 32×32 pixel views and a memory footprint under 9 kB, our system offers a biologically grounded, resource-efficient solution for autonomous visual homing.

\end{abstract}

\section{Introduction}

Navigation in unstructured environments remains a key challenge in robotics. Desert ants (such as \textit{Cataglyphis} or \textit{Melophorus}) excel at this using coarse vision-based strategies such as obstacle avoidance \cite{franceschinietal.InsectVisionRobot1992}, path integration (PI) \cite{seeligNeuralDynamicsLandmark2015}, route-following \cite{kohlerIdiosyncraticRoutebasedMemories2005}, and homing \cite{zeilVisualHomingInsect2012,collettNeuroethologyAntNavigation2025},despite their small brains and sparse visual encoding \cite{wystrachHowFieldView2016}. These capabilities have inspired insect-based navigation models in robotics \cite{dupeyrouxAntBotSixleggedWalking2019,manganVirtuousCycleInvertebrate2023}.

This study focuses on visual homing—returning to a known location via visual cues, ideally along a direct path (beeline) \cite{gaussierVisualHomingProblem2000,pfefferAccuracySpreadNest2020}. Prior robotic approaches, not using panoramic views, have incorporated PI through multiple sensors \cite{dupeyrouxAntBotSixleggedWalking2019}, and Central Complex (CX)-inspired computational models \cite{seeligNeuralDynamicsLandmark2015,stankiewiczUsingNeuralCircuit2020,stoneAnatomicallyConstrainedModel2017}. However, purely path-integration-based methods are prone to drift due to cumulative sensor errors. Hence, visual-based strategies, including Fourier transform matching and catchment area gradient descent\cite{mollerLocalVisualHoming2006,stoneRotationInvariantVisual2018,vandijkVisualRouteFollowing2024}, have emerged as promising alternatives, though their biological grounding remains uncertain.

While many robotic methods use images taken at the goal location as the only homing reference \cite{caron2013photometric,lambrinosMobileRobotEmploying2000}, studies in ants show that individuals can learn the broader vicinity of the nest via meandering exploratory behaviors, known as learning walks \cite{zeil2025coming,pfefferAccuracySpreadNest2020}. It remains an open research question how PI signal from the CX interact with visual memory encoding in  Mushroom Bodies (MBs) \cite{ardinUsingInsectMushroom2016,mullerPathIntegrationProvides2010,wystrachNeuronsPremotorAreas2023}. While MB-inspired models have been applied to robotic route following \cite{gattaux2023antcar,gattaux2024continuous}, visual homing with MBs has so far only been explored in simulation \cite{wystrachNeuronsPremotorAreas2023}. A recent neurobiological model proposes a lateralized MB architecture, where the angular sign of the PI output from the CX guides visual learning during learning walks by categorizing views as “goal on the left” or “goal on the right” relative to heading direction \cite{wystrachNeuronsPremotorAreas2023}.

Building on the lateralization principle \cite{wystrachNeuronsPremotorAreas2023}, we have developed and demonstrated how a MB-based model can achieve robust robotic visual homing outdoor. Our system categorizes panoramic views during learning walks into three memory banks type based on information from the PI : goal on the left, goal on the right, or at the goal location, to perform visual homing under natural outdoor conditions. We validate this system through four incremental experiments: (1) simulation in a 3D world showing attractor dynamics toward the nest; (2) real-world outdoor homing using left/right memory banks inferred from two decoupled circular learning walks, eliminating the need for PI; (3) homing after random learning walks using a PI emulated with a GPS-RTK (Global Positioning System - Real-Time Kinematic) as a noisy learning scaffold, enabling homing and search behavior up to 18 meters from the nest; and (4) precise docking behavior using a fifth Nest MB Output Neuron (MBON) that modulates velocity based on nest familiarity.

Our results demonstrate that robust visual homing can emerge in real-world environments using coarse visual input, minimal memory, and low-cost sensors. Highlighting the value of integrating angular PI signals into visual memory systems, offering a lightweight and biologically grounded framework for autonomous visual homing.

\section{Methods}

\subsection{Mobile robot's description}

\begin{figure}[b!]
\centering
\includegraphics[width=0.9\textwidth]{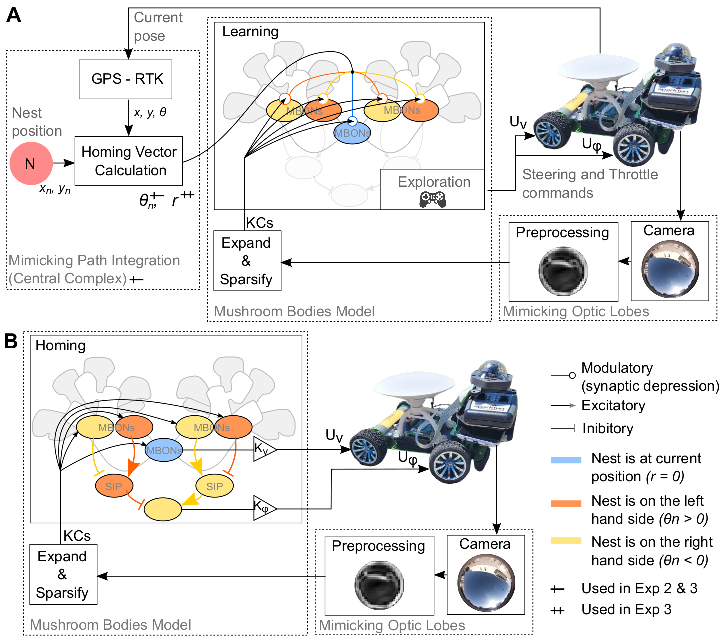}
\caption{Embedded processing pipeline in the AntCar robot. \textbf{(A)} \textit{Learning phase}: Images captured by the onboard camera are processed through modules mimicking insect optic lobes and Mushroom Bodies (MB), facilitating visual learning and homing. GPS-RTK provides positional information used for computing path integration vectors during learning, simulating input from the insect Central Complex. The learning phase is initiated through human teleoperation. The MB model comprises five Mushroom Body Output Neurons (MBONs). During learning, visual information encoded by Kenyon Cells (KCs) is associated with MBONs modulated by a homing vector signal, acting analogously to dopaminergic feedback neurons (DANs). \textbf{(B)} \textit{Homing phase}: During homing, KC activity is compared with previously stored MBON patterns; differences among four MBONs determine steering ($U_\varphi$), while familiarity encoded by the fifth MBON controls throttle ($U_v$).}
\label{figmeth1}
\end{figure}

Experiments were performed using the AntCar robot \cite{gattaux2023antcar}, a compact (25 cm length) car-like robot with DC-powered rear wheels and servo-controlled front steering. Processing was fully onboard, managed by a Raspberry Pi 4B due to its efficiency and low power consumption. Motors were controlled via an I2C-connected PCA9685 module. Power was provided by three rechargeable 18650 batteries (2600 mAh each, total 12.6 V).

Outdoor localization used a SparkFun GPS-RTK Surveyor module, providing positional accuracy up to 14 mm at 2 Hz. Ground speed and heading were computed by differentiating positional data. During learning phases, the mobile robot was teleoperated via joystick for controlled exploration around the nest (Fig. \ref{figmeth1}A).

Visual input was obtained using an Entaniya\texttrademark~ fisheye camera, mounted atop the robot, capturing a 220° vertical and 360° horizontal field of view at 30 Hz. All processing modules (Fig. \ref{figmeth1}) were implemented as ROS Python nodes.

\subsection{Embedded Processing Modules}
 
Embedded processing involved three interconnected processing stages (Fig. \ref{figmeth1}A): optic lobes, PI (functional mimicking), and MBs.

\subsubsection{Mimicking Optic Lobes}
The captured images were processed to mimic the ant's visual system. Images were first filtered with a Gaussian low-pass filter to reduce noise, downsampled to thumbnails of 32×32 pixels (equivalent to approximately 7°/pixel resolution), and then processed using a Sobel high-pass filter to highlight edges. The resulting filtered images were flattened into a vector termed Projection Neurons (PN).

\subsubsection{Mimicking Path Integration}

The PI module computed the global homing vector from GPS-RTK data and a predefined nest location. GPS latitude and longitude coordinates were transformed into Cartesian coordinates using Lambert projection. The nest position was encoded egocentrically as the angular orientation relative to the robot’s heading ($\theta_n$) and the Euclidean distance ($r$). GPS heading accuracy was continuously monitored. This processing stage is used in \rotatebox{90}{\dag} Exp. 2 and 3 only. 

\subsubsection{The Mushroom Bodies Model}

The MBs model, structured for visual memory encoding and retrieval, was adapted from previous bio-inspired approaches \cite{ardinUsingInsectMushroom2016,gattaux2023antcar} with lateralized modifications inspired by recent neurobiological models \cite{wystrachNeuronsPremotorAreas2023}. The model involved two distinct phases: learning and homing (Fig. \ref{figmeth1}B):

\begin{itemize}
    \item \textbf{Expansion and Sparsification}: Projection Neurons (PNs, 750 neurons) were expanded into 5,000 Kenyon Cells (KCs) through a semi-random, fixed synaptic weight matrix (PNtoKC). To achieve sparse representation, a k-Winner-Take-All mechanism retained only the 50 most active KCs (top 1\%) for encoding visual information.
    
    \item \textbf{Learning Phase}: Visual scenes encoded by KCs were selectively associated with one of five Mushroom Body Output Neurons (MBONs), guided by the PI vector obtained via GPS-RTK. Novel visual inputs triggered synaptic depression in the designated synaptic weight vector (KCtoMBON), dynamically reducing synaptic weights unit from an initial value of 1 (unlearned) to 0 (learned). Specifically, scenes observed with the nest on the robot’s left were learned by the "left-side" MBON (orange), scenes with the nest on the right by the "right-side" MBON (yellow), and scenes observed directly at the nest by a dedicated fifth MBON (blue).
    
    \item \textbf{Homing Phase (Autonomous Navigation)}: During homing, incoming KC patterns were compared against stored MBON representations, generating familiarity indices from 0 (familiar) to 1 (unfamiliar). During homing, familiarity was dynamically assessed by calculating a weighted sum between the current KC activation pattern and the learned KC-to-MBON synaptic weights, followed by normalization by the total number of active KCs (1\% of total neurons). Steering commands resulted from differential familiarity between the left and right MBONs within each Mushroom Body hemisphere and were further integrated and differentiated in the Superior Intermediate Protocerebrum (SIP), scaled by a fixed gain factor. Additionally, the robot's speed was continuously modulated by the fifth MBON's familiarity, enabling the robot to precisely stop at the familiar nest location without explicit conditional logic. This MBON processing stage is used in \rotatebox{90}{\ddag} Exp. 3 only.
\end{itemize}

\subsection{Experimental Protocol}

\begin{figure}[t!]
\centering
\includegraphics[width=0.9\textwidth]{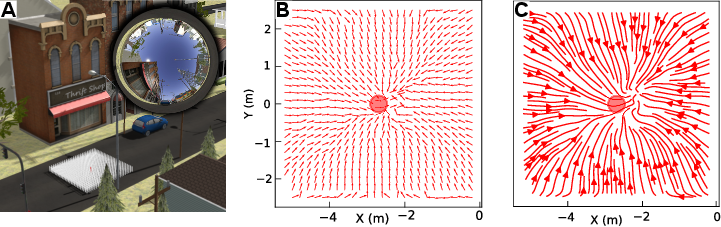}
\caption{Estimated homing direction based on visual familiarity after learning with integrated panoramic views and path integration (PI) in a simulated urban environment. The home location is indicated by a red circle. \textbf{(A)} Simulated Gazebo environment showing the grid of sampled positions (white squares) where panoramic images were captured for learning and testing. Inset: example camera viewpoint. \textbf{(B)} Homing directions estimated using visual familiarity, trained with binary left/right outputs from a PI system. \textbf{(C)} Stream plot depicting possible continuous homing trajectories.}
\label{figres1}
\end{figure}

The initial experiments were conducted in a simulated environment—a small city built in Gazebo—using a virtual fisheye camera matching the real robot’s visual system. The simulated world was systematically sampled over a 5×5 m grid, containing 673 unique spatial positions ($x$, $y$) (Fig. \ref{figres1}A). At each grid location, panoramic images were captured at orientations separated by 10°, resulting in a total dataset of 24,228 learned and tested images.

For both the simulation and initial outdoor experiments (Experiments 1 and 2), only the four lateralized MBONs were employed (without velocity modulation), therefore the robot maintained its speed at about 0.8 m/s. Three distinct nest positions were chosen (latitude, longitude) for the outdoor experiments. Learning walks consisted of trajectories around each nest location to encode visual memories from multiple directions. In Experiment 1, the two learning walks were performed separately with the nest positioned either to the left or right of the robot (decoupled walks, without using the PI module in Fig. \ref{figmeth1}). In Experiment 2, the GPS-RTK integration allowed randomized (coupled) learning walks without directional constraints. Experiment 3 (\rotatebox{90}{\ddag}, Fig. \ref{figmeth1}) introduced a velocity modulation controlled by an additional MBON, enabling precise stopping behavior at the nest.

\section{Results}

A demonstration video is provided with this study\footnote{Video available at: https://youtu.be/2pZ1nTxyf2k}. It includes preliminary tests conducted in Toulouse (unpublished), as well as the real-world experiments described in the results section. During every experiments, the robots was kidnapped and placed randomly at different start locations (crosses at Fig. \ref{figres2},\ref{figres3} and \ref{figres4})

\begin{table}[b!]
\centering
\renewcommand{\arraystretch}{1.5}  % tighter vertical spacing
\setlength{\tabcolsep}{8pt}        % tighter horizontal spacing
\caption{Average angular error (in degrees) between estimated (visual) and ground truth homing angles in simulation, with corresponding control update rates (Hz) on a Raspberry Pi 4 using GPS-RTK.}
\label{tab1}
\begin{tabular}{c|c|c}
\hline
Parameters & r = 7 (°/px) & r = 5 (°/px)\\
\hline
N = 5,000 & $19.78 \pm 21.44$° (9 Hz) & $20.75 \pm 23.64$° (7.5 Hz) \\
N = 10,000 & $16.03 \pm 19.51$° (6.9 Hz) & $15.29 \pm 17.84$° (5.2 Hz) \\
N = 50,000 & $11.14 \pm 9.62$° (3 Hz) & $10.01 \pm 10.17$° (1.7 Hz) \\
\hline
\end{tabular}
\end{table}

\subsubsection{Simulation Results: Attractor Dynamics and Homing Accuracy}

The attractor-like dynamics of the visual homing strategy were demonstrated through the bearing map derived from familiarity differences in the SIP (Fig. \ref{figres1}B). A stream plot further illustrates convergence toward the nest position (Fig. \ref{figres1}C). To quantify homing accuracy, we compared the angular error between the MB model’s estimated homing angle and the ground truth (Table \ref{tab1}). The best absolute mean angular error (\(10.01 \pm 10.17^\circ\)) was achieved with a network size of \(N = 50,000\) neurons and a visual resolution of \(r = 5^\circ/pixel\). 

However, increasing visual resolution yields only marginal performance gains while significantly raising computation time (Table \ref{tab1}). For this reason, and to offer a practical trade-off between efficiency and robustness, outdoor experiments were conducted with \(N = 5,000\) neurons and \(r = 7^\circ/pixel\). This configuration improved the control update frequency by a factor of five compared to the highest-performance setting.

\subsubsection{Experiment 1: Orbiting learning walks with Nest Left/Right} 

\begin{figure}[b!]
\centering
\includegraphics[width=0.9\textwidth]{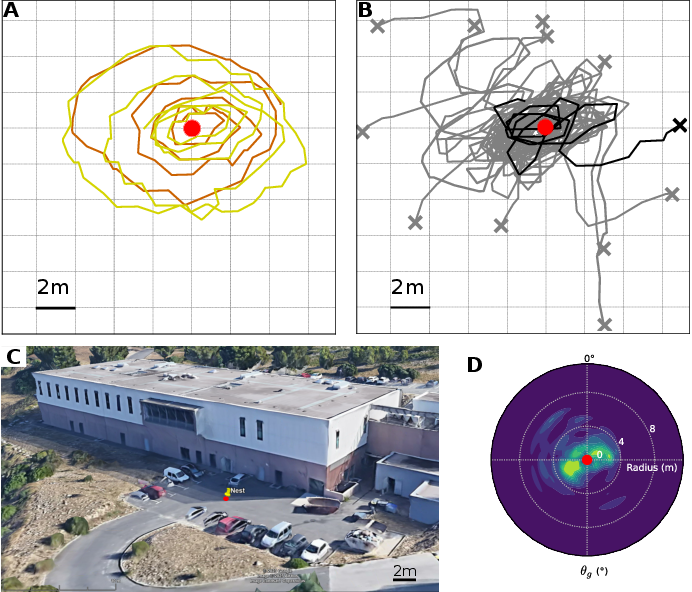}
\caption{Experiment 1: in Marseille, France, demonstrating online visual learning and homing. \textbf{(A)} Decoupled learning walks: views stored in left (orange) and right (yellow) MBONs during separate clockwise and counterclockwise orbits around the nest. \textbf{(B)} Autonomous homing: the robot returns and searches near the nest from various kidnapped and replaced positions (crosses). \textbf{(C)} Google Earth overview showing nest location (GPS coordinates: 43.234535° N, 5.443509° E). \textbf{(D)} Catching area: polar distribution of robot positions during homing.}
\label{figres2}
\end{figure}

In Experiment 1, the robot conducted two separate circular learning walks around the nest located in Marseille, France at (43.234535° N, 5.443509° E) (Fig. \ref{figres2}C). The counterclockwise  walk (nest on the left) covered a total distance of approximately 75 meters, learned 1,234 views, and had a maximum radius of 6.18 meters. The clockwise walk (nest on the right) covered approximately 103 meters, learned 1,401 views, and reached a maximum radius of 6.70 meters. Visual inputs were learned at a rate of approximately 14 Hz during these "learning walks".

During autonomous homing tests (Fig. \ref{figres2}B, D), control commands were updated at approximately 13.5 Hz. The robot was tested from 12 distinct locations surrounding the nest, at distances up to 12 meters—about twice the radius of the learning walks. In each trial, the robot successfully returned and continuously searched within a limited radius of the nest (mean search distance \(2.4 \pm 1.8\) m), confirming a robust visual homing behavior.

\subsubsection{Experiment 2: Enhanced Learning with a noisy PI}

\begin{figure}[b!]
\centering
\includegraphics[width=0.9\textwidth]{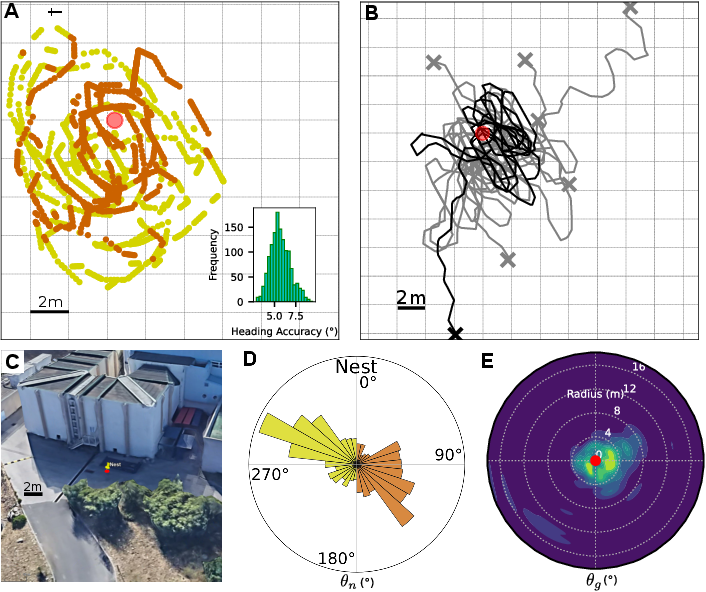}
\caption{Experiment 2: in Marseille, France, demonstrating online visual learning and homing with GPS-RTK integrated for global PI. \textbf{(A)}  Learning walk: views stored separately in left (orange) or right (yellow) MBONs based on nest direction; inset shows GPS heading accuracy. \textbf{(B)} Homing: robot successfully orbits nest after displacement from multiple positions. \textbf{(C)} Google Earth overview indicating nest location (43.234245° N,, 5.443668° E). \textbf{(D)} Distribution of views learned (left/right). \textbf{(E)} Polar plot distribution of robot positions during homing.}
\label{figres3}
\end{figure}

In Experiment 2, we used GPS-RTK to mimick a noisy PI and derive its angular sign (binary) left/right output to supervise visual learning in the MB. This enables the robot to learn along less structured, literally randomized learning walks (Fig. \ref{figres3}A, D). The nest was located in Marseille, France at (43.234245° N, 5.443668° E) (Fig. \ref{figres3}C). The learning walks covered approximately 282 meters, with 740 views learned (399 left-side and 341 right-side views, Fig. \ref{figres3}D), reaching a maximum radius of 8.17 meters. Visual inputs were learned at about 8 Hz, less than the experiments 1 likely due to the GPS integration (info at 2Hz) and PI calculation. 

After learning, the memory footprint (synaptic weights learned vector stored as Compressed Sparse Rows, CSR) was only 8.74 kilobytes. Due to real-world conditions, heading accuracy provided by GPS-RTK averaged about \(5^\circ\) error (Fig. \ref{figres3}A inset), testing the robustness of visual learning against noisy PI data.

During homing (Fig. \ref{figres3}B, E), control commands were computed at approximately 7 Hz. The robot successfully homed from 7 distinct locations around the nest, positioned up to 18 meters away—more than twice the radius of the learning walks. In each trial, the robot returned and searched consistently near the nest (mean search distance: \(3.4 \pm 3\) m, Fig. \ref{figres3}E). With a slightly increased mean search distance and variance compared to Experiment 1 (Fig. \ref{figres2}D).

These findings demonstrate robust visual homing performance despite inherent noise and limitations of real-world navigation conditions.

\subsubsection{Experiment 3: Precise Homing Using Noisy PI and Additional Nest MBON}

\begin{figure}[b!]
\centering
\includegraphics[width=0.9\textwidth]{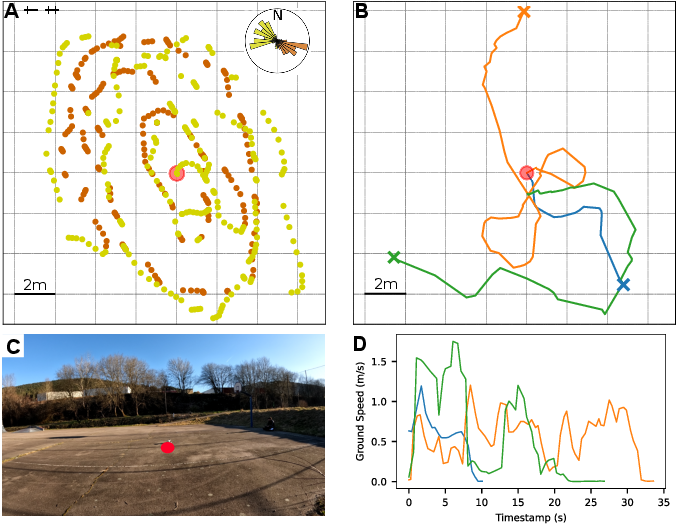}
\caption{Experiments 3: Gerardmer, France, demonstrating online visual learning using GPS-RTK PI and precise homing using an additional MBON controlling velocity. \textbf{(A)}  Learning walk: views stored in left (orange), right (yellow), or nest-specific MBON (all directions). Inset shows distribution of learned views. \textbf{(B)} Precise homing task: robot accurately returns to and stops at the nest from three positions. \textbf{(C)} Photo overview of environment (48.079883°N, 6.888538°E). \textbf{(D)} Robot's ground speed over time.}
\label{figres4}
\end{figure}

Visually homing ants can sometimes miss their nest entrance by a centimeter, and thus engage in a visually driven local search around the nest location,
\cite{zeil2025coming,schultheiss2013information} similarly to our robot in the previous experiments. Some ant species further refine nest pinpointing using olfactory cues when in close proximity \cite{steck2009smells}. To mimic this stopping behavior, we extended our model with an additional MBON dedicated to recognizing views directly over the nest and modulating the robot's velocity. In addition to categorizing panoramic views into left and right MBONs during learning walks (as in the previous experiment; Fig. \ref{figres4}A), this new MBON encoded views captured while the robot was located at the nest (Fig. \ref{figmeth1}, $r = 0$). The experiment was conducted in an open-field site in Gerardmer, France (48.079883° N, 6.888538° E), with scattered trees and no buildings (Fig. \ref{figres4}C). Learning walks covered approximately 163 meters and included 811 learned views: 348 left-side, 463 right-side, and 360 nest-centered views (captured during an initial full 360° rotation). The robot learned visual inputs at $\sim 8$ Hz, consistent with previous experiments. The learned synaptic weights (KC-to-MBON connections) were stored in Compressed Sparse Row (CSR) format, achieving a memory footprint reduction of approximately 99.89\% compared to raw storage, decreasing from 7.8 MB to 8.85 kB.

During homing (Fig. \ref{figres4}B) the robot successfully returned from three distinct locations around the nest, up to 8 meters away—matching the radius of the learning walks. In each trial, it stopped precisely at the nest (0.02, 0 and 1.04 m). The controlled velocity decreased as the familiarity signal from the Nest MBON decreased, then dropped near the nest (Fig. \ref{figres4}D). These stopping behaviors highlight the potential of this approach for precise robotic docking, closely resembling a standard position-controlled task. The addition of a velocity-controlling MBON significantly improves homing accuracy and closely mirrors natural ant docking behavior.

\section{Discussion}

In this study, we demonstrated robust visual homing using a lightweight, bio-inspired neural network suitable for low-resource robotic platforms. Our approach integrated panoramic visual cues into a lateralized MB model with five MBONs, modulated by the angular, binary sign of the PI output during learning. The model proved resilient to very low visual resolution and noisy PI data (GPS inaccuracies), achieving stable homing despite intermittent PI updates (2 Hz) relative to visual updates (8 Hz).

Our work complements recent convolutional neural network approaches that learn to associate positions based on odometric data with visual cues for homing in simulated drones \cite{firlefynDirectLearningHome2024}. Rather than learning to associate each view to a specific coordinate, our model categorizes views into only three categories ($2\times$left, $2\times$right, and $1\times$nest) enabling efficient operation at 8 Hz on a Raspberry Pi 4, using only 32×32 pixel inputs and less than 9 kB of memory. This highlights its suitability for lightweight, resource-constrained robotic platforms, and reflects the minimalist yet robust navigation strategies employed by insects.

The robot's behavior observed during Experiments 1 and 2 resembled the natural visually-driven searching behavior in ants when missing their nest entrances \cite{pfefferAccuracySpreadNest2020,zeil2025coming,schultheiss2013information}. However, differences in robot kinematics influence search mechanics compared to those observed in ants (notably the limited maximum steering angle of the robot). By introducing an additional MBON dedicated to recognise the scenes perceived from the nest location itself (Experiment 3), the robot could precisely halt at the nest site, mimicking the accurate homing behavior ants display. We successfully approximated this precision with visual memories alone \cite{zeil2025coming}, reinforcing the versatility of our MB model. 

Interestingly, Table \ref{tab1} reveals that with fewer neurons (N = 5,000), the model performs slightly better at lower visual resolution (7°/pixel) compared to higher resolution (5°/pixel), whereas at higher neuron counts (N = 10,000 and 50,000), performance improves at higher resolution (5°/pixel). This result indicates that the current approach, i.e activating a fixed percentage (1\%) of KCs, may not be universally optimal. Indeed, fewer active KCs at lower neuron counts appear beneficial for forming generalized visual representations at coarser resolutions. To enhance robustness and efficiency across varying environments and resolutions, future models could benefit from activating a fixed absolute number of KCs, chosen independently from the total number of neurons and more aligned with environmental complexity and desired visual accuracy.

Future work should also optimize the patterns of learning walks to enhance visual homing effectiveness. Additionally, employing insect-inspired CX models with on board ego-motion sensors for PI could eliminate GPS reliance, facilitating seamless indoor/outdoor operation \cite{stankiewiczUsingNeuralCircuit2020}. In addition, as shown in ants, the CX could be equally used during tests, as a relay between the MB outputs and motor commands to ground visual guidance on compass cues, which adds resilience to noisy visual recognition \cite{sun2019modelling,wystrach2020lateralised,wystrachNeuronsPremotorAreas2023}. A more in-depth analysis of the model’s robustness to visual complexity and environmental modifications—such as altered landmarks or background clutter—could help define the system's operational limits and inform improvements in generalization and adaptability. Finally, integrating optic lobe neural models and coupling our homing approach with route-following strategies \cite{gattaux2024continuous} would further enhance biological plausibility and practical navigation capabilities. 

\vspace{5mm}
\begin{credits}
\noindent
\textbf{Code Availability Statement.} The code is available upon request. A video of the experiments is available at https://youtu.be/2pZ1nTxyf2k .

\subsubsection{\ackname} We would like to thank Léo Clement and Hamidou Diallo for their help with the experiments. G.G. was supported by a doctoral fellowship grant from Aix Marseille University (amU) and the French Ministry of Defense (AID - Agence Innovation D\'{e}fense, agreement \#A01D22020549 ARM/DGA /AID). G.G., J.R.S. and F.R. were also supported by Aix Marseille University and the CNRS Institutes (Biology, Informatics as well as Engineering). The facilities for the experimental tests has been mainly provided by ROBOTEX 2.0 (Grants ROBOTEX ANR-10-EQPX-44-01 and TIRREX ANR-21-ESRE-0015).

\subsubsection{\discintname}
All authors declare that they have no conflicts of interest.

\end{credits}
%
% ---- Bibliography ----
%
% BibTeX users should specify bibliography style 'splncs04'.
% References will then be sorted and formatted in the correct style.
%
\bibliographystyle{splncs04}
\bibliography{mybibliography}

\end{document}